\documentclass[sigconf]{acmart}

\usepackage{graphicx}
\usepackage{multicol}
\usepackage[utf8]{inputenc}
\usepackage[ruled,vlined]{algorithm2e}
\usepackage{amsmath}
\usepackage{algpseudocode}
\usepackage{float} 
\DeclareMathOperator*{\argmax}{arg\,max}

\RestyleAlgo{ruled}
\SetKwComment{Comment}{/* }{ */}
\usepackage{tikz}
\usepackage{comment}
\usepackage{mathtools}
\usepackage{amsfonts}
\usepackage{enumitem}
\usepackage{xcolor,listings}
\usepackage{color,soul}
\definecolor{backcolour}{rgb}{0.95,0.95,0.92}
\sethlcolor{backcolour}
\usepackage{textcomp}
\newcommand*{\code}[1]{\hl{\texttt{#1}}}
\newcommand*{\class}[1]{\emph{#1}}

\newcommand*\circled[1]{\tikz[baseline=(char.base)]{
            \node[shape=circle,draw,inner sep=1pt] (char) {#1};}}

\begin{document}

\title{A Framework for History-Aware Hyperparameter Optimisation in Reinforcement Learning}

\author[Juan Marcelo Parra-Ullauri, Chen Zhen, et al.]{Juan Marcelo Parra-Ullauri \and Chen Zhen \and Antonio García-Domínguez \and Nelly Bencomo \and Changgang Zheng \and Juan Boubeta-Puig \and Guadalupe Ortiz \and Shufan Yang }
\begin{abstract}
A Reinforcement Learning (RL) system depends on a set of initial conditions (\emph{hyperparameters}) that affect the system's performance. However, defining a good choice of hyperparameters is a challenging problem. 
Hyperparameter tuning often requires manual or automated searches to find optimal values. Nonetheless, a noticeable limitation is the high cost of algorithm evaluation for complex models, making the tuning process computationally expensive and time-consuming.
    In this paper, we propose a framework based on integrating complex event processing and temporal models, to alleviate these trade-offs. Through this combination, it is possible to gain insights about a running RL system efficiently and unobtrusively based on data stream monitoring and to create abstract representations that allow reasoning about the historical behaviour of the RL system. The obtained knowledge is exploited to provide feedback to the RL system for optimising its hyperparameters while making effective use of parallel resources.
    We introduce a novel \emph{history-aware epsilon-greedy logic} for hyperparameter optimisation that instead of using static hyperparameters that are kept fixed for the whole training, adjusts the hyperparameters at runtime based on the analysis of the agent's performance over time windows in a \emph{single agent's lifetime}.  
    We tested the proposed approach in a 5G mobile communications case study that uses DQN, a variant of RL, for its decision-making. Our experiments demonstrated the effects of hyperparameter tuning using history on training stability and reward values. The encouraging results show that the proposed history-aware framework significantly improved performance compared to traditional hyperparameter tuning approaches.
\end{abstract}
\maketitle 
\section{Introduction}
\label{intro}
Reinforcement Learning (RL) is a sub-field of Machine Learning with a great success in applications such as self-driving cars, industry automation, among many others~\cite{0c}. In RL, autonomous agents learn through trial-and-error how to find optimal solutions to a problem~\cite{0c}. RL algorithms have multiple \emph{hyperparameters} that require careful tuning as it is a core aspect of obtaining the state-of-the-art performance~\cite{jomaa2019hyp}. 

The search for the best hyperparameter configuration is a sequential decision process in which initial values are set, and later adjusted, through a mixture of intuition and trial-and-error, to optimise an observed performance to maximise the accuracy or minimise the loss~\cite{jomaa2019hyp}. Hyperparameter Optimisation (HPO) often requires expensive manual or automated hyperparameter searches in order to perform properly on an application domain~\cite{zahavy2020self}. However, a noticeable limitation is the high cost related to algorithm evaluation, which makes the tuning process highly inefficient, computational expensive, and commonly adds extra algorithm developing overheads to the RL agent decision-making processes~\cite{zhang2021importance,zahavy2020self,jomaa2019hyp,feurer2019hyperparameter}.

The full behaviour of complex RL systems often only emerges during operation. They thus need to be monitored at runtime to check that they adhere to their requirements~\cite{rabiser2017comparison}. Event-driven Monitoring (EDM) is a common lightweight approach for monitoring a running system~\cite{klar1992tools}. Particularly, \emph{Complex Event Processing} (CEP) is an EDM technique, for capturing, analysing, and correlating large amounts of data in real time in a domain-agnostic way~\cite{luckham1998complex}. The present paper proposes the use of CEP to quickly detect causal dependencies between events on the fly by continuously querying data streams produced by the RL system in order to gain insights from events as they occur during the execution of the RL agent which is crucial for HPO~\cite{feurer2019hyperparameter}. 

CEP provides the short-term memory needed to analyse the system behaviour on pre-defined time-points or limited time-windows. However, it is debated that long-term memory is also required when analysing the effects of HPO on the RL agent to find optimal performance evolved on past behaviours. \emph{History-awareness} requires node-level memory and traceability management facilities to allow the exploration of system's history. \emph{Temporal Models} (TMs) are seen to tackle these challenges~\cite{gomez2018temporalemf}. TMs offer storage facilities that allows time representation using a temporal database (TDB)~\cite{parra-ullauri_event-driven_2021, parra2021towards}. In this paper, a TDB supports the storage of massive amounts of historical data, while providing fast querying capabilities to support reasoning about runtime properties in the monitored RL agent.

In this paper, we propose a framework based on CEP ans TMs that can be reused for different RL algorithms. The proposed combination allows the detection of situations of interest at runtime and permits tracing the RL agent history to enable the short and long term memory required to analyse the impact of HPO. The framework uses a formal defined structure to trace data streams produced by the RL agents, process them and provide feedback for HPO. In addition, we present a novel \emph{history-aware epsilon-greedy logic} for HPO that is implemented using the components of the proposed framework. This logic tunes the hyperparameter concurrently while acting greedily under certain circumstances, but also exploring the hyperparameter value-space with an $\epsilon$ probability in order to escape local maximums. The HPO occurs while the agent is learning, which turns to be more efficient than using static hyperparameters during the training process and having to update them on multiple agent's lifetimes~\cite{zahavy2020self,zhang2021importance}.

In order to test the feasibility of the proposed framework, Deep Q-Network (DQN)~\cite{0c}, a popular RL algorithm, was applied to a case study on the next generation of mobile communications from~\cite{zheng2021reward}. The experiments analysed the effects of the proposed history-aware approach for HPO during the RL agent training, and compared the results with traditional hyperparameter tuning approaches. Our experiments focused on updating the discounting factor hyperparameter at runtime for a \emph{single agent's lifetime}, using these different techniques. 

The rest of the paper is organised as follows. Section~\ref{background} provides a description of the core concepts required to understand this paper. Section~\ref{proposal} introduces our approach. Experiments and results are presented in Section~\ref{caseStudy}. The discussion is presented in Section~\ref{discussion}. Section~\ref{relatedWork} compares the presented work with current state of HPO in RL. Finally, Section~\ref{conclusion} presents conclusions and future directions.

\section{Background}
\label{background} 
\subsection{Hyperparameter Optimisation in Reinforcement Learning}
In RL, an agent tries to maximise the optimal action-value function described as the Bellman Optimality Eq.~\cite{0c}:
\vspace{-0.4 em}
\begin{equation}
    Q^*(s, a)= \mathbb{E}\{r_{t+1}+\gamma*max_{a'}Q^*(s_{t+1}, a')|s_{t}=s, a_{t}=a\}
    \label{eq:bellman}
\vspace{-0.5 em}
\end{equation}
where $\mathbb{E}$ represents the expected sum of future rewards characterised by the hyperparameter $\gamma$, which is the \emph{discounting factor}~\cite{0c}.
A reward $r_t$ that occurs N steps in the future from the current state, is multiplied by $\gamma^N$ to describe its importance to the current state. As shown, defining the right $\gamma$ and additional hyperparameters through HPO is key to deliver optimal solutions in RL. 

The most basic way of HPO is manual search, which is based on the intuition of the developer~\cite{feurer2019hyperparameter}. Once the system execution has finished, the verification of convergence is reviewed. More sophisticated HPO approaches include i) Model-free Blackbox Optimisation (MBO) and ii) Bayesian Optimisation (BO)~\cite{feurer2019hyperparameter}. Grid and random search are part of i). In grid search, the user defines a set of hyperparameter values to be analysed and the search evaluates the Cartesian product of these sets. Random search samples configurations at random until a certain budget for the search is exhausted.~\cite{feurer2019hyperparameter}. Regarding ii), BO iteratively evaluates a promising hyperparameter configuration based on the current model then updates it trying to locate the optimum in multiple agent's lifetimes. However, performing these techniques is time consuming, computationally expensive and requires expert knowledge~\cite{fernandez2018parameters}.

For the reasons mentioned, the introduction of an automated hyperparameter search process is key for the continuing success of RL and is acknowledged as the most basic task in automated machine learning (AutoML)~\cite{feurer2019hyperparameter}. In this work we focus on MBO for a \emph{single agent’s lifetime}, which is claimed to be more efficient than having static hyperparameters during the training process and updating them in multiple agent's lifetimes~\cite{zahavy2020self,zhang2021importance}.

\subsection{Temporal Models}

TMs go beyond representing and processing the current state of systems~\cite{gomez2018temporalemf}. They seek to add short and long-term memory to models through the use of temporal databases ~\cite{mazak2020temporal}. Examples of temporal databases used for TMs, are Time Series Databases (TSDB) and Temporal Graph Databases (TGDB)~\cite{mazak2020temporal,parra-ullauri_event-driven_2021}. Each attribute to be monitored in a running system can be considered as a time series: a sequence of values along an axis~\cite{esling_time-series_2012}. TGDB extend this ability to track the appearance and disappearance of entities and connections~\cite{greycat}.

TGDBs record how nodes and edges appear, disappear and change their key/value pairs over time. Greycat~\cite{greycat} is an open-source TGDB. Nodes and edges in Greycat have a lifespan: they are created at a certain time-point, they may change in state over the various time-points, and they may be ``ended'' at another time-point. Greycat considers edges to be part of the state of their source and target nodes. It also uses a copy-on-write mechanism to store only the parts of a graph that changed at a certain time-point, thus saving disk space. In this work, TMs build on top of Greycat TGDB, allow accessing and retrieving causally connected historical information about runtime behaviour of RL agents.

\subsection{Event-driven Monitoring}
EDM approaches are commonly designed to monitor system events, processes and handle them in the background without interfering with the main system's execution~\cite{moser2010event}. Moser et al.\ identified in \cite{moser2010event} four key requirements for EDM: i) it should be platform agnostic and unobtrusive, ii) it should be capable of integrating monitoring data from other subsystems, iii) it should enable monitoring across multiple services and instances, and iv) it should be capable of unveiling potential anomalies in the monitored system. CEP is a cutting-edge EDM technology that has been widely used to address these requirements~\cite{luckham1998complex}.

CEP provides real-time analysis and correlation of large volumes of streaming data in an effective and efficient manner with the aim of automatically detecting situations of interest in a particular domain (event patterns). The patterns to be detected have to be defined and deployed into a CEP engine, i.e.\ the software responsible for analysing and correlating the data streams. Each CEP engine provides its own Event Processing Language (EPL) for implementing the patterns to be deployed. Among the existing CEP engines, we opted for Esper\footnote{https://www.espertech.com/esper/}, a mature, scalable and high-per\-for\-mance CEP engine. The Esper EPL is a language similar to SQL but extended with temporal, causal and pattern operators, as well as data windows. The present document proposes to leverage the power  of CEP to detect temporal and causal dependencies between events and to pre-process data streams, in order to gain insights from events as they occur during the training of an RL agent.

\section{History-Aware Hyperparameter Optimisation for RL}
\label{proposal}
This section presents our proposed framework integrating CEP and TMs for HPO. Additionally, the section also describes a novel history-aware epsilon-greedy logic that will be implemented using the referenced framework.

\subsection{A Software Framework combining CEP and TMs for HPO}
\label{sec:framework}

\begin{figure}
    \centering
    \includegraphics[width=\columnwidth]{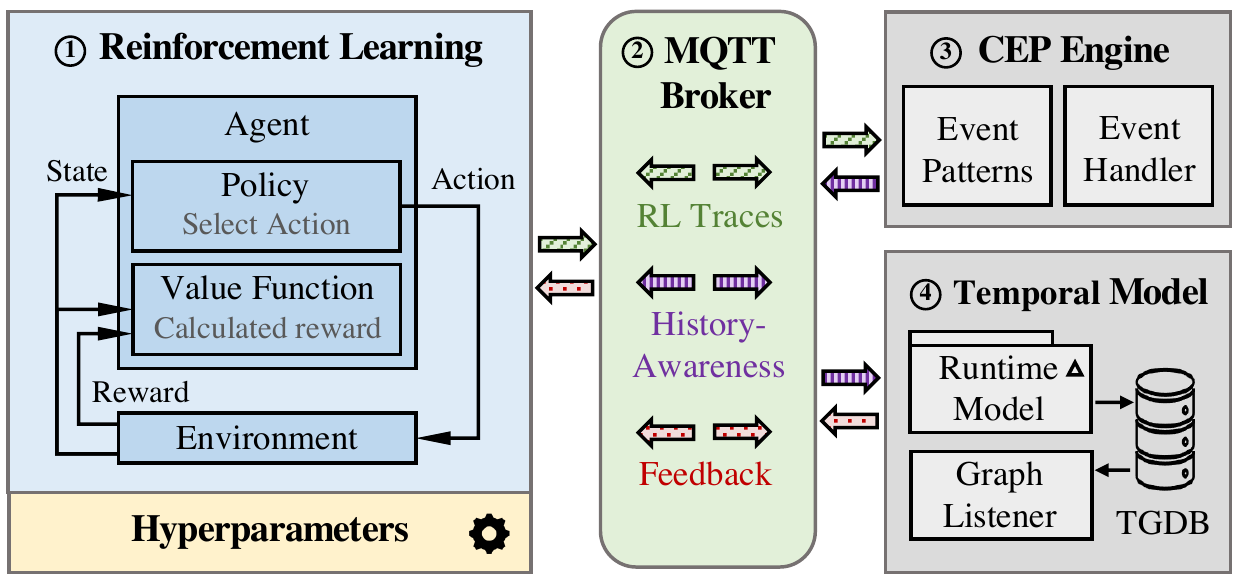}
    \caption{CEP and TMs for Hyperparameter Optimisation}
    \label{fig:cep_tms}
\end{figure}

RL involves challenging optimisation problems due to the stochasticity of evaluation, high computational cost and possible non-stationarity of the hyperparameters~\cite{zhang2021importance}. Therefore, the efficient continuous monitoring and dynamic verification of internal operations and parameters of the RL agent and its interactions with the environment over time are required. We propose the use of CEP for short-term analysis and TMs for navigation through the system history to provide feedback to the RL algorithm. Fig.~\ref{fig:cep_tms}
 shows the proposed framework:

\begin{itemize}
\item The \emph{RL algorithm} \circled{1} runs mostly independently from the rest of the system, while publishing data streams with logging information into an ``RL Traces'' topic created in a Message Queuing Telemetry Transport (MQTT) broker. The algorithm is subscribed to a ``Feedback'' topic, which will contain suggestions for hyperparameter change.

\item The \emph{MQTT Broker} \circled{2} is the communication hub for the architecture, acting as an event bus. It is responsible for loosely integrating the other components through the use of \emph{topics}: components can publish events into a topic, or subscribe to updates about that topic.

\item The \emph{CEP Engine} \circled{3} is responsible for filtering and correlating data streams in the form of simple events coming from the RL algorithm into semantically richer~\emph{complex events}. It subscribes to the ``RL Traces'' topic to obtain those simple events, and it pushes complex events into the ``History-Awareness'' topic.

\item The \emph{Temporal Model} \circled{4} uses the complex events from the ``History-Awareness'' topic to construct the next version of the high-level model of the RL agent's state, which is used to update the TM. A novel \emph{graph listener} component is notified about the changes, which applies the HPO logic (see Section~\ref{sec:hpo-logic}) to push any feedback on the current hyperparameter values into the ``Feedback'' MQTT topic.
\end{itemize}

\begin{figure}
    \centering
    \includegraphics[width=\columnwidth]{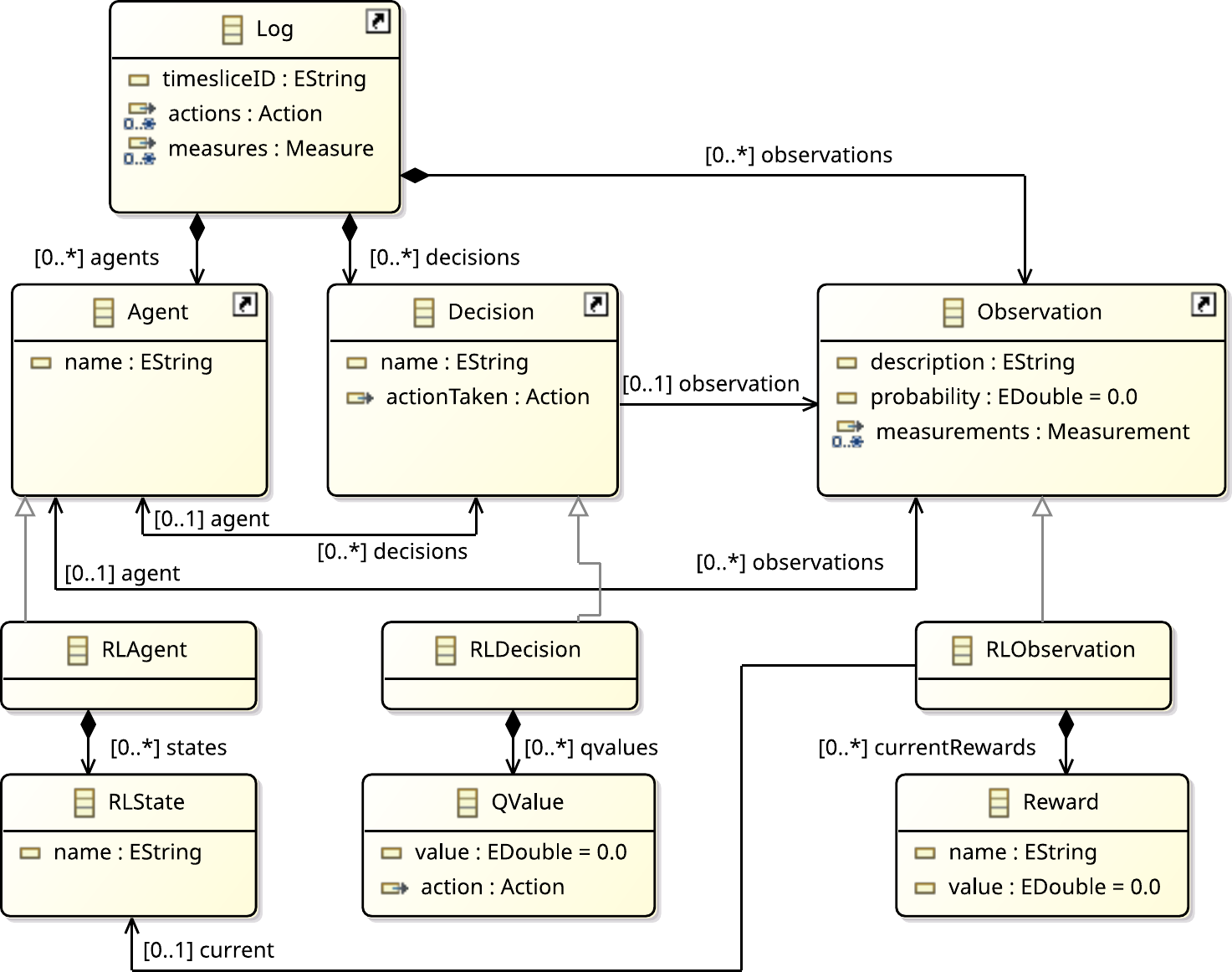}
    \caption{Class diagram for the RL extensions to the core metamodel used to record system history. Imported core elements are marked with an arrow.}
    \label{fig:mm-rl}
\vspace{-0.5em}
\end{figure}

TMs are conceptually structured according to a metamodel designed to record a \class{Log} of \class{Decision}s made by \class{Agent}s, based on \class{Observation}s about the environment, and including a set of \class{Measurement}s of interest, according to various \class{Measure}s. The metamodel is divided into two parts: the above concepts are defined into a core package from [\emph{omitted}], and concepts that are specific to RL are split into its own package (see  Fig.~\ref{fig:mm-rl}), which imports elements from the core package. The RL package provides a specialised \class{RLAgent} which keeps track of the \class{RLState} that can be observed in the environment, an \class{RLDecision} which tracks the \class{QValue}s of each available action, and an \class{RLObservation} which tracks the current state before the action was taken, and the current \class{Reward} values.

\subsection{History-aware epsilon-greedy logic for HPO}
\label{sec:hpo-logic}

In RL, the $N$-dimensional hyperparameter configuration space is defined as $\Lambda$~=~$\Lambda_1 \times ... \times \Lambda_N$ and a vector of hyperparameters is denoted by $\lambda \in \Lambda$. Let's denote the RL algorithm as $\Phi$ and $\Phi_\lambda$ the algorithm instantiated to a vector of hyperparameters $\lambda$. Let us define the objective function to maximise the value of a reward function $\mathcal{R}$. Then, we define the HPO problem of a $\Phi$ given the environment $E$ at time $T$ as finding the optimal hyperparameter vector $\lambda^*$:
\begin{equation}
    \lambda^*=\argmax_{\lambda \in \Lambda}\mathcal{R}(\Phi_\lambda , E , T)
    \label{eq:hpo_problem}
\end{equation}

where $\mathcal{R}(\Phi_\lambda , E , T)$ measures a reward value generated by the algorithm $\Phi$ under a configuration of the $\lambda$ hyperparameter while interacting with the environment $E$ at time $T$.

Now we can introduce our history-aware epsilon-greedy approach. RL is episodic, with multiple iterations $i \in I$ performed within each episode $e \in E$~\cite{0c}. In this context, let us define the value of $\mathcal{R}$ at the instant $t_{i}$ after $\Phi_\lambda$ has interacted with the environment $E$ as the reward $r_t$ that the agent obtained by performing an action $a_t$ and arriving to the state $s_t$. Thus, the value of our reward function by iteration is denoted by $\mathcal{R}_{i}(t_i,\lambda)=r_{t_i}$. Consequently, the reward function by episode is defined by:
\begin{equation}
    \mathcal{R}_{e}(t_e,\lambda)=\frac{\sum_{i=1}^{I} \mathcal{R}_{i}(t_i,\lambda)}{I} 
    \label{eq:rewardbyepisode}
\end{equation}
After stated our reward function by episode, we define the criterion for analysing the history. In other words, how long back are we going to look when deciding to change a hyperparameter. With this purpose, we introduce the concept of time-windows to the logic. A time-window $w \in W$ consists of $x \in \mathbb{R}$ episodes $e$ where $x$ is the length of the time-window. Then, the value of our reward function by time-window is denoted by:

\begin{equation}
    \mathcal{R}_{win}(t_w,\lambda)=\frac{\sum_{w=e}^{e+x-1} \mathcal{R}_{e}(t_w,\lambda)}{x}
    \label{eq:rewardbywindow}
\end{equation}

Eq.~\ref{eq:rewardbywindow} defines the time frame when the monitoring process is taking place. The next step is to define the criterion that would lead to a hyperparameter update. 
The criteria selected is the \emph{stability} of the reward value. We analyse the distance of the reward function by episode $\mathcal{R}_e$ to the mean of the time-window $\mathcal{R}_{win}$. If the value is below a defined threshold $th_{stable} \in \mathbb{R}$ for all the values within the time-window, we induce that the reward value has stabilised within a range and a \emph{possible} hyperparameter update will be performed. Formalising this as a Boolean conjunction we have:
\begin{equation}
   \label{eqn:stability}
    \displaystyle \bigwedge_{j \in [e, e + x)} \left(|\mathcal{R}_{e}(t_j,\lambda)-\mathcal{R}_{win}(t_w,\lambda)|<th_{stable}\right)
\end{equation}
where $\mathcal{R}_{win}$ is the reward function value for the time window at time $w=e+x-1$.

We have emphasised the word \emph{possible} for a change in $\lambda$, as stability won't necessarily mean that the agent has reached its maximum performance under the current conditions. Let's consider the example when our optimiser system has observed the following set of $\mathcal{R}_e$ under the same conditions $\Phi_\lambda$, $\mathcal{R}:\{1,2,3,4,5,6\}$. We define a time-window length of 3 episodes ($x=3$) and a stability threshold of 2 ($th_{stable}=2$). Thus, $w_1=\{1,2,3\}$, $w_2=\{4,5,6\}$, $\mathcal{R}_{win_{1}}=2$ and $\mathcal{R}_{win_{2}}=5$. As a result, the Boolean conjunction for $w_1$ will be true. This would mean that the system requires a hyperparameter change. However, under the same conditions $\Phi_\lambda$, the system would have kept improving its performance as it is shown in $w_2$ and $\mathcal{R}_{win_{2}}$. Therefore, an additional condition is necessary to define when a hyperparameter tuning is required.

We introduce $max\mathcal{R}_t$ as the maximum known value of $\mathcal{R}_{win}$ up to the time-point $t$ and it is initialised as 0. Similarly, we introduce $max\lambda_t$ as the value of $\lambda$ that has produced $max\mathcal{R}_t$ up to the time-point $t$. We then define the main condition for hyperparameter tuning $HPO(\lambda)$ and it is described as follows:  
\begin{equation}
    HPO(\lambda)= 
\begin{cases}
   \quad \begin{rcases}
        \lambda_t, &\\ 
        \quad \text{\textbf{with}} &\\
        max\lambda_t \gets \lambda_t &\\
        max\mathcal{R}_t \gets \mathcal{R}_{win_{t}}
    \quad \end{rcases}& \text{if } \mathcal{R}_{win_{t}} > max\mathcal{R}_t\\
    \\
   \quad \begin{rcases}
    \xi(\lambda), \qquad \qquad \qquad
   \quad \end{rcases}& \text{otherwise}
\end{cases}
\label{eqn:hop_lambda}
\end{equation}

where $HPO(\lambda)$ is equal to $\lambda_t$ \texttt{iff} the current value of $\mathcal{R}_{win_{t}}$ is greater than the maximum known value of $max\mathcal{R}_t$ at time $t$. This would imply that the current value of our $\mathcal{R}$ function has increased from the previous maximum known and therefore the current configuration $\Phi_{\lambda_{t}}$ should be kept as the system is still `learning'. Correspondingly, the current value of $\mathcal{R}_{win_{t}}$ would become the new maximum known value of $max\mathcal{R}_t$ ($max\mathcal{R}_t \gets \mathcal{R}_{win_{t}}$). In the case that the previous condition for $HPO(\lambda)$ is not met ($\mathcal{R}_{win_{t}} > max\mathcal{R}_t$), a hyperparameter tuning is required and will be analysed by our epsilon-greedy function $\xi(\lambda)$. 

Our optimiser would examine $\xi(\lambda)$ \texttt{iff} and only \texttt{iff} the following conditions are met: i) $\mathcal{R}$ is stable for a time-window $w$ (Eq. 5), and ii) the current value of $\mathcal{R}_{win_{t}}$ is less than the previous maximum known value of $\mathcal{R}$, $max\mathcal{R}_t$. These conditions mean that the system is stable (regarding to rewards observed) and on a sub-optimal configuration of $\Phi_{\lambda_{t}}$ (on reference to $max\mathcal{R}_t$). Therefore, a different vector of hyperparameters $\lambda$ should be explored. The criterion for selecting the next $\lambda$ is a variation of the well-known epsilon ($\epsilon$)-greedy policy for balancing exploration and exploitation in RL~\cite{0c}. The optimiser will explore the hyperparameter value-space with a probability of $\epsilon$, otherwise it will exploit the known best configuration ($max\lambda_t$). Thus, it would act greedily. Eq. 8 shows the proposed epsilon-greedy function $\xi(\lambda)$.
\begin{equation}
    \xi(\lambda)=
    \begin{cases}
        \quad\begin{rcases}
        max\lambda_t, \qquad \quad 
        \quad\end{rcases}& \text{with probability 1-$\epsilon$}\\
\\
        \quad\begin{rcases}
        \text{random } \lambda \in \Lambda, &\\
        \quad \text{\textbf{or}} &\\
        \lambda + c, &\\
        \quad \text{\textbf{or}} &\\
        \lambda - c, 
        \quad\end{rcases}& \text{with probability $\epsilon$}
    \end{cases}
\label{eqn:hop_epsilon}
\end{equation}

In addition of exploring randomly the value-space with a probability $\epsilon$, we have introduced supplementary conditions that will help the optimiser for deciding if the current value of the hyperparameter to which it is performing the tuning should be increased to $\lambda +c$ or decreased to $\lambda -c$, where $c$ is a user-defined constant. These conditions give hints to the optimiser about the direction in the value-space where better performance is achieved.

Putting everything together, given a RL algorithm $\Phi$ that interacts with an environment $E$ at time $t \in T$ with an initial hyperparameter configuration $\lambda \in \Lambda$ such as $\Phi_\lambda$, its configuration will be analysed and possibly updated based on $HPO(\lambda)$ when meeting the stability condition on a time-window from (\ref{eqn:stability}). The update based on our $\epsilon$-greedy function $\xi(\lambda)$ will only take place if the observed value of our $\mathcal{R}$ for such a time-window is less than the best known value of $\mathcal{R}$, $max\mathcal{R}$. The advantage of the proposed approach is that exploration actions are only selected in situations when the system has stopped learning under the defined conditions, which is indicated by analysing the history of $\mathcal{R}$.

Finally, the found optimal hyperparameter vector $\lambda^*$ for a lifetime of the RL agent, corresponds to the final value of $max\lambda$:
\begin{equation}
    \lambda^* \gets max\lambda
\end{equation}

This history-aware epsilon-greedy logic for HPO in RL has been implemented using the architecture in Section~\ref{sec:framework}, exploiting the benefits of CEP and TMs.

\section{Experiments and Results}
\label{caseStudy}
\subsection{System under study: Airbone base stations}
In order to demonstrate the feasibility of the proposed architecture, this section presents its implementation for a case study from the domain of mobile communications. In this case study, Airborne Base Stations (ABS) use DQN to decide where to move autonomously in order to provide connectivity to as many users as possible. The 5G Communications System Model performs the necessary calculations to estimate the Signal-to-Interference-plus-Noise Ratio (SINR) and the Reference Signal Received Power (RSRP) to defined if a user is connected or not~\cite{zheng2021reward}.

Developed by DeepMind in 2015, DQN has produced some breakthrough applications able to solve a wide range of Atari games even more efficiently than humans~\cite{mnih2015human}. In contrast to tabular RL approaches, DQN avoids using a lookup table by instead predicting the Q-value of the current or potential states and actions using artificial neural networks (NN) or deep learning networks~\cite{0c}. This Q-function (see Eq.~\ref{eq:bellman}) provides the expected discounted reward that results from taking an action $a_t$ in the state $s_t$ and policy $\pi$ is followed.
The hyperparameters when training a DQN agent include the numbers of episodes and neurons, learning rate, exploration rate, discount factor, among others. 

\subsection{Experimentation: Scenario}
For the current implementation, we decided to study the impact of optimising the discount factor as the key element in the Bellman equation after a non-exhaustive manual hyperparameter search. The discount factor determines how much the RL agents cares about rewards in the distant future relative to those in the immediate future~\cite{0c}.
The hyperparameter vector is expressed as follows:
\begin{equation*}
    \lambda(\gamma,\kappa)
\end{equation*}
where $\gamma$ represents the hyperparameter to be optimised (i.e. discount factor) and~$\kappa$ the hyperparameters that remain fixed.

With these preliminaries, different experiments using the proposed framework were performed under the same scenario. It consisted of a training round (i.e. a single lifetime) of 100 episodes and 1000 steps for a set of 4 ABS with 1050 users scattered on a X-Y plane. The ABS try to maximise the number of users connected by performing actions (i.e. moving on different directions) and calculating the SINR to users on a collaborative fashion. Our main goal was trying to solve Eq.~\ref{eq:hpo_problem}. With this purpose, two different experiments were defined:

\begin{itemize}
    \item \textbf{History-Aware HPO vs traditional HPO}: 
In order to evaluate the proposed approach, we benchmarked it with traditional HPO techniques. Specifically, the previous mentioned (see Section~\ref{background}), grid search and random search and BO. For the seek of the experiment, the hyperparameter tuning was performed uusing Optuna hyperpameter tool~\cite{akiba2019optuna} using these HPO techniques and during the training process. Thus during a single agent's lifetime different of the common use of these approaches that requires the analysis on multiple agent's lifetimes~\cite{zahavy2020self}. In this context, the initial hyperparameters for each approach is described in table~\ref{table_experiment1}. From the literature~\cite{0c,mnih2015human}, commonly used values for the discount factor are within the range of 0.9 and 0.99. We have included the manual setting with static discounting factor for comparison. For the case of grid search, in order to cover the hyperparameter-value space, the initial value of $\gamma$ is set to 0.9 and decays over the time with a rate of 0.1. Random search starts randomly, the history-aware HPO and BO start in the centre of the value-space.  

\item \textbf{History-Aware HPO vs static hyperparameters}: 
A second experiment was performed to analyse the impact of our approach in the performance of the system in comparison to keeping the hyperparameters static during the training. For this purpose, we initialised the training round with the same seeds and compared the reward evolution overtime for each hyperparameter configuration. The initial values of the discount factor that conformed the experiment were: $\gamma_{o} \in \{0.1,0.2,0.3,0.4,0.5,0.6,0.7,0.8,0.9\}$.

\end{itemize}

\begin{table}
\centering
\begin{tabular}{||c c c||} 
 \hline
 Approach &  $\gamma_{o}$ & Tuning criteria \\ [0.5ex] 
 \hline\hline
 Manual setting & 0.9 & static \\
 Grid search  & 0.9 & updated every 10 episodes \\ 
 Random search & random & updated every 10 episodes \\
 Bayesian Optimisation & 0.5 & updated every 10 episodes \\
 History-aware HPO & 0.5 & automated-tuning \\ [1ex] 
 \hline
\end{tabular}
\caption{Initial configuration for experiment 1}
\label{table_experiment1}
	\vspace{-2em}
\end{table}

\subsection{Experimentation: Setup}
In order to test the feasibility of our approach, the history-aware epsilon-greedy logic for HPO presented in Section~\ref{sec:hpo-logic} was implemented using the different components of the proposed framework of Section~\ref{sec:framework}. Accordingly, the process depicted in Fig.~\ref{fig:cep_tms} is described next:
\begin{enumerate}
    \item RL algorithm: DQN has been the selected RL approach for the system under study. The algorithm was extended to send the made decisions and observations in a trace log to a queue in a MQTT message broker in JSON format at each simulation step.
    \item MQTT broker: The open-source Mosquitto\footnote{https://mosquitto.org/} was selected as communication hub. The different components are subscribed to topics that allow them to send and receive messages on the network using a publish/subscribe model.
       \item CEP engine: It processes and correlates the trace logs received from the RL algorithm with the aim of detecting, in real time, the situations of interest for the application domain. A set of event patterns were implemented in the selected CEP engine (Esper). Precisely, we have implemented Equations 4, 5 and 6 using Esper EPL in a hierarchy of event patterns. Listing~\ref{lst:timeWindow} shows the implementation of Equation 6 that attempts to detect stable conditions on time windows of 3 episodes ($w=3$) with a $th_{stable}=30$. Every pattern \code{AvgByEpisode} (which refers to Eq.~\ref{eq:rewardbyepisode}), followed (\code{->}) by two subsequent \code{AvgByEpisode} and a \code{EpiWinAVG} (which refers to Eq.~\ref{eq:rewardbywindow}), is analysed (\code{where} statement) in compliance of the boolean conjunction of Eq.~\ref{eqn:stability}. When the condition is met (i.e. boolean conjunction = $True$), the engine automatically generates complex events that collect the required information, to therefore send the events to the MQTT broker component for further processing.
    \item Temporal Model: It receives complex events and records their information as a new version of the model in the TGDB. Specifically, the Hawk\footnote{https://www.eclipse.org/hawk/} model indexer was extended with the capability to subscribe to an MQTT queue and reshape the information into a model conforming to the metamodel in Fig.~\ref{fig:mm-rl}. The graph listener is notified when a stable condition is detected. Next, it performs the required calculations of Eq.~\ref{eqn:hop_lambda} and~\ref{eqn:hop_epsilon} to provide feedback to the RL algorithm on either; keeping the hyperparameter configuration or tuning it towards finding a good solution to Eq.~\ref{eq:hpo_problem}. The feedback provided is recorded as part of the temporal model enabling the long-term memory needed for further processing, accountability and post-mortem analysis.  
    
\end{enumerate}

\begin{lstlisting}[
  float,language=SQL,
  caption={Esper EPL pattern to select when the system is on an stable condition on a defined time-window},
  label=lst:timeWindow,
  numbers=none,backgroundcolor=\color{backcolour},
  columns=flexible,
  breakatwhitespace=false,         
    breaklines=true,
    numbersep=5pt,                  
    showspaces=false,                
    showstringspaces=false,
    showtabs=false,                  
    tabsize=2
]
@public @buseventtype @Name("isStableAVG")
insert into isStableAVG
select w.averageWindow as avg, 
       w.episode as episode, 
       'CEP' as agent, 
       CAST(a3.drone_number as INT) as drone_number,
       a3.step as step, 
       a3.gamma as gamma
from pattern [every a1 = AvgByEpisode -> 
                    a2 = AvgByEpisode -> 
                    a3 = AvgByEpisode -> 
                    w = EpiWinAVG] 
where 
  w.episode = a3.episode and 
  Math.abs(a1.avg-w.averageWindow) < 30 and
  Math.abs(a2.avg-w.averageWindow) < 30 and
  Math.abs(a3.avg-w.averageWindow) < 30
\end{lstlisting}

As previously mentioned, our implementation decouples the running RL-system from the HPO process. In that sense, the experiments were performed using two machines dedicated to different purposes: one performing the training of the different RL algorithms, and the other running the proposed framework. The RL algorithms ran on a dedicated ML server with 10 NVIDIA RTX A6000 48GB GPUs using the ABS simulator, Python 3, Anaconda 4.8.5, matplotlib 3.3.4, numpy 1.19.1, paho-mqtt 1.5.0, pandas 1.1.3, and pytorch 1.7.1. The machine running the proposed framework was a Lenovo Thinkpad T480 with an Intel i7-8550U CPU with 1.80GHz, running Ubuntu 18.04.2 LTS and Oracle Java 1.8.0\_201, using Paho MQTT 1.2.2, Eclipse Hawk 2.0.0, and Esper 8.0.0. The full implementation can be found in (omitted for double-blind review).

\subsection{Evaluation of the results}
In this section, we present the evaluation of the results of using the proposed framework implementing the history-aware epsilon-greedy logic for HPO. We trained the DQN system under the same conditions for the different experiments. A total of 20 runs were conducted.

\subsubsection{History-Aware HPO vs traditional HPO}
\begin{figure}
    \centering
    \includegraphics[width =\columnwidth]{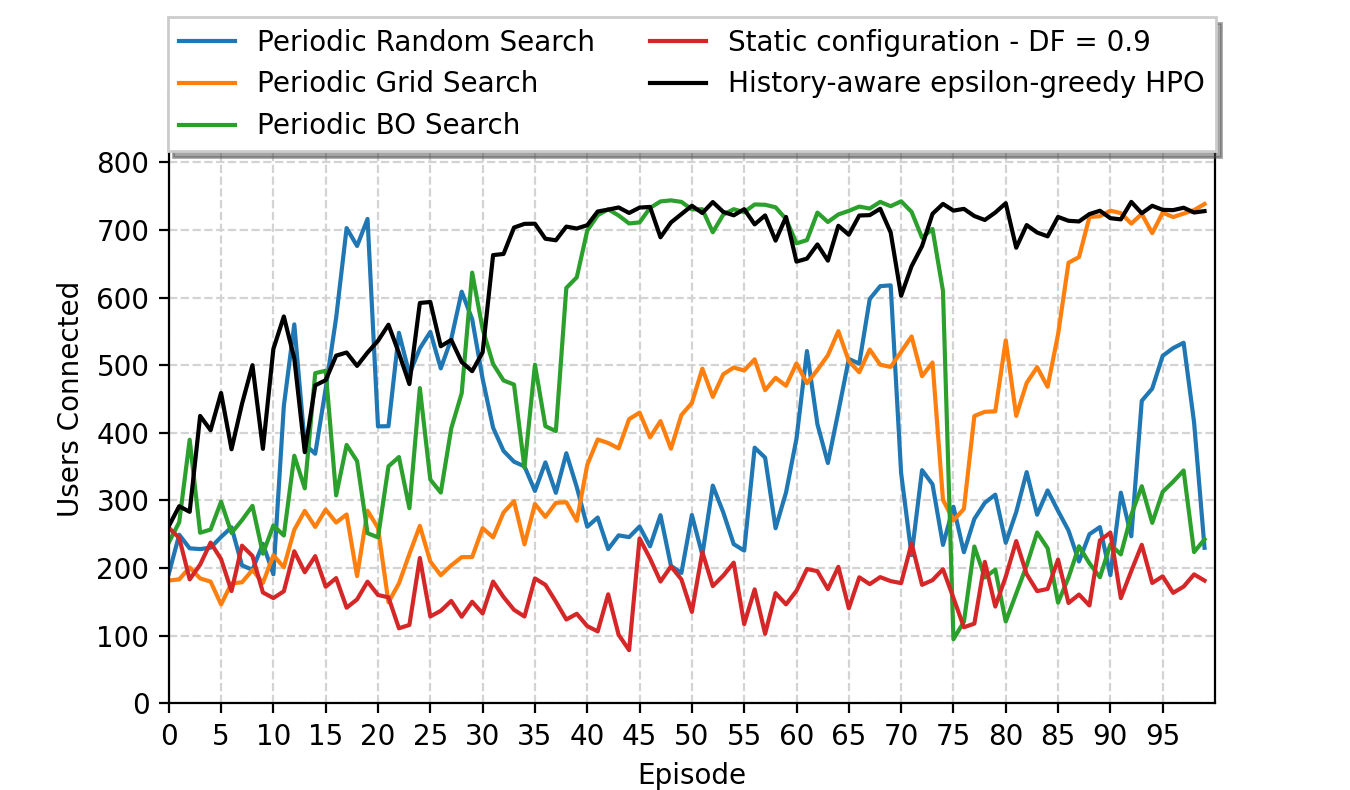}
    \caption{Comparison of hyperparameter tuning methods in DQN}
    \vspace{-1em}
    \label{fig:dqn_comparison}
\end{figure}

The first experiment corresponded to a qualitative study of the performance of the ABS system using the proposed approach comparing against traditional HPO techniques. Fig.~\ref{fig:dqn_comparison} shows the results. As it can be observed, our history-aware HPO approach (black line) over-performed, in terms of time to converge and accuracy, the different approaches obtaining its maximum values from episode 32 onward. 
The random search (blue line) fluctuates and its performance is closed with static configuration (red line). It is interesting to note that grid search (green line) achieved similar performance. However, the sharp dip at episode 70 to 80 shows a potential instability. Similarly, BO (green line) achieves maximum performance in episode 42 however, it could not recover after trails of sub-optimal hyperparameter values from episode 71 onwards.    

\begin{figure}[t]
	\centering
	\begin{minipage}{0.45\linewidth}
	\includegraphics[width=1\linewidth]{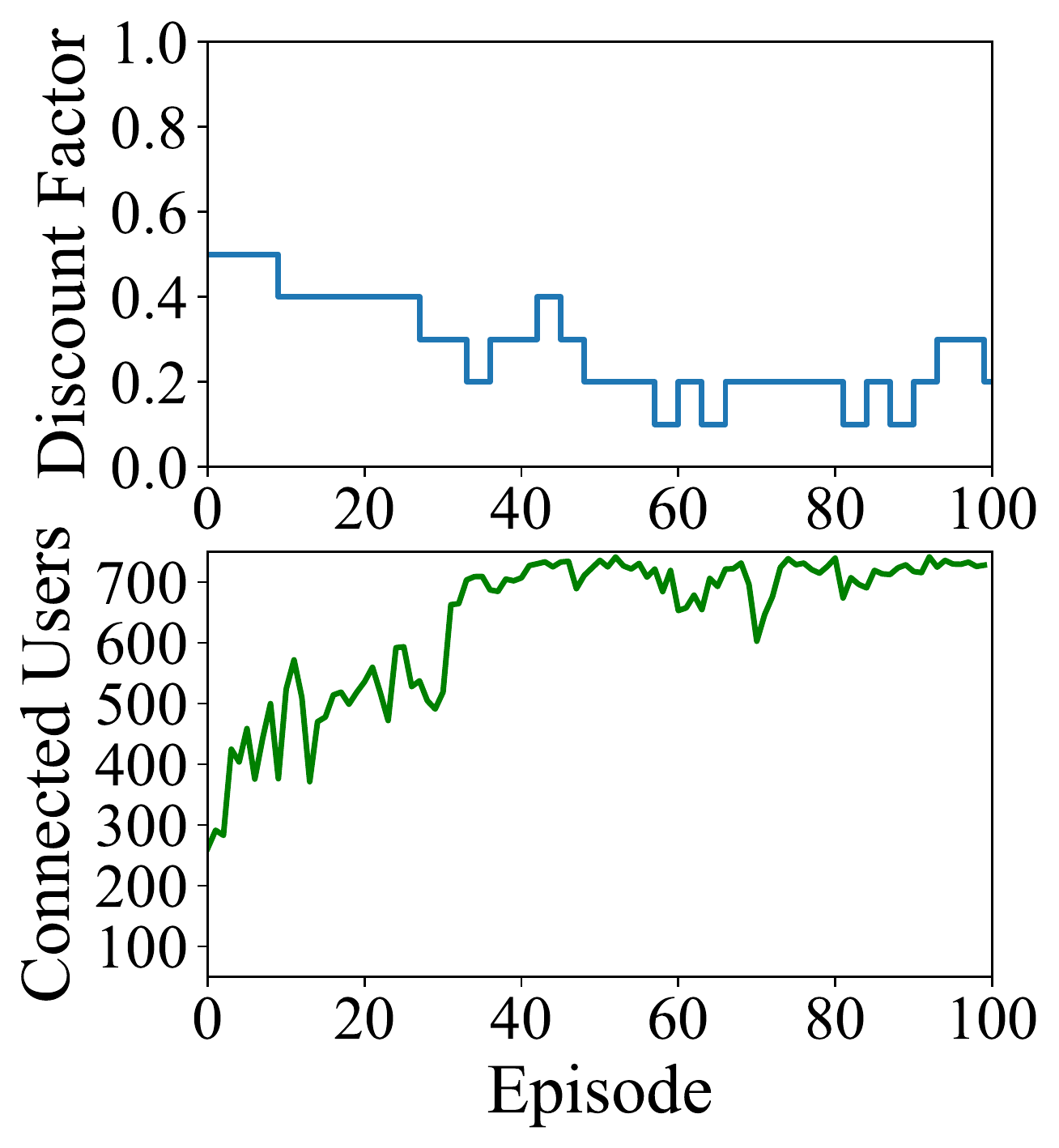}\vspace{-0.5em}\\\centering(a) Gamma = 0.5.
\end{minipage}
	\begin{minipage}{0.45\linewidth}
	\includegraphics[width=1\linewidth]{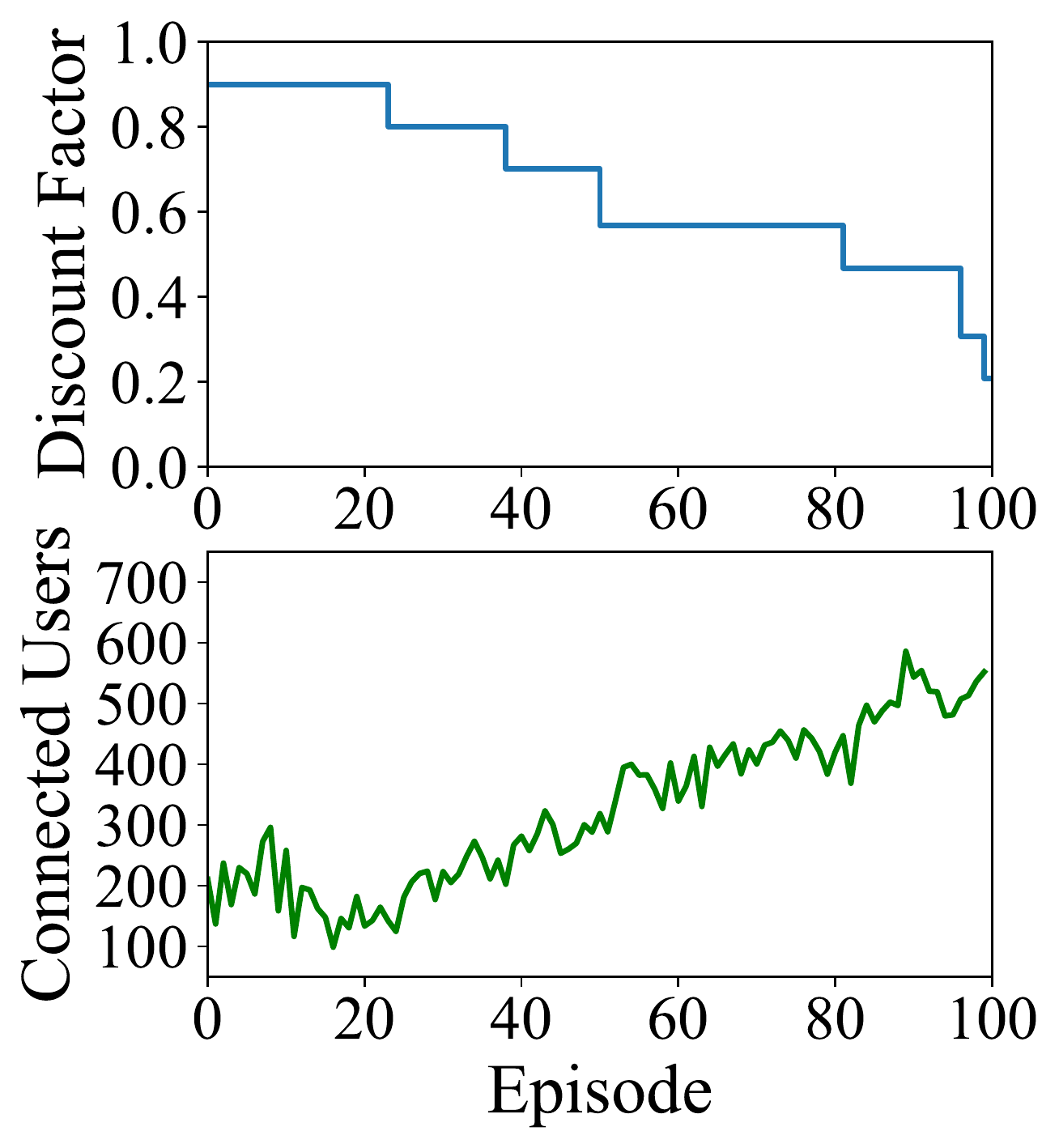}\vspace{-0.5em}\\\centering(b) Gamma = 0.9.
\end{minipage}
	\vspace{-0.5em}
	\caption{Reward and discount factor evolution, starting at \textbf{$\gamma =0.5$} and \textbf{$\gamma =0.9$} using history-aware HPO.}\label{fig:gamma}
	\vspace{-2em}
\end{figure}

The proposed approach allows us to get more insights about the HPO process by analysing the history stored in the TGDB in conformation of metamodel of~\ref{fig:mm-rl}. Fig.~\ref{fig:gamma}~(a) depicts the results. The extracted information shows that the maximum value of our reward value function $\mathcal{R}$ was 727.055 at episode 74 with $\gamma$=0.204. Therefore, under the configuration $\Phi_{\lambda(\gamma,\kappa)}$ the optimal found value for the HPO problem of Eq. 3 is: $\lambda^* \gets \lambda(\gamma=0.204,\kappa)$.

\subsubsection{History-Aware HPO vs static hyperparameters}
\begin{figure}
    \centering
    \includegraphics[width =0.8\columnwidth, height=130pt]{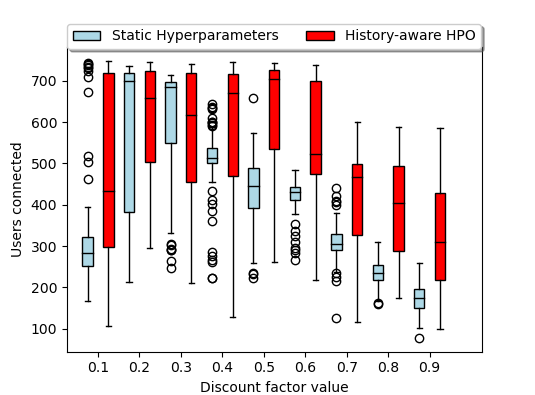}
    \caption{Comparison of history-aware hyperparameter optimisation vs static values}
    \label{fig:dqn_comparison_static}
\end{figure}

The second experiment included an exhaustive analysis of the performance of the RL algorithm using different seeds for the discount factor. The comparison include the analysis of the reward value function $\mathcal{R}$ with and without the proposed approach for each system configuration $\Phi_{\lambda_{t}(\gamma,\kappa)}$. The boxplots of Fig.~\ref{fig:dqn_comparison_static} display the results. By using the proposed history-aware HPO (in red) the system was able to reach greater maximum values (the upper end of the whiskers) for each configuration. Furthermore, the iterquartile ranges (boxes) in each case had a greater upper quartile. Regarding to the medians, that represent the middle of the set, they were also greater for each case except for $\gamma = 0.2$ and $\gamma = 0.3$. This can suggest two things: i) the optimal values of $\gamma$ is within this range $0.2<\gamma^*<0.3$, which reinforce the result obtained in experiment 1, and ii) the variance in the data corresponds to the optimiser exploring the hyperparameter value-space with probability $\epsilon$. The results showed that by the use of the approach no outliers that lie on an abnormal distance from other values in the data set were found.  

The best performance of the RL system using the history-aware HPO approach occurred when the initial value of the discounting factor was the centre of the hyperparameter value-space, $\gamma_{o}=0.5$ with an average of connected users of 636.104 and a median of 702.886. In the same manner, the poorest performance occurred with $\gamma_{o}=0.9$ with an average of 309.818 connected users and a median of 323.774. As shown in Fig.~\ref{fig:gamma}~(b), after exploring the hyperparameter value-space, the optimiser was going towards the optimal value of $\gamma$ which is corresponded with the increase of the reward. Thus, the system would have needed longer to find the optimal value.

\section{Discussion}
\label{discussion}

The results from our conducted experiments showed the feasibility of the history-aware approach for HPO. Combining CEP and TMs made us to offer both the short and long term memory required for hyperparameter tuning with reflective capabilities. The history-aware epsilon-greedy logic allowed to explore the hyperparameter value-space with explicit long-term memory to remember good/optimal system's configurations $\Phi_{\lambda_{t}(\gamma,\kappa)}$. Our experiments provide valuable insights into the effects of the tuning of the discount factor and its influence on the stability of training and overall system  performance (maximised cumulative rewards).

The discount factor determines how many future time steps the agent considers when choosing an action. This value strongly depends on the environment that software agents are experienced. In the ABS case study, a discount factor close to 1 allows the agent to take actions very future oriented. A lower discount factor suggests that the ABS are more concerned to  provide coverage to multiple users in short term but would introduce uncertainty in the long term. It is challenging to find the balance between the highest possible number of connected users in the short term and the long-term impact, as user behaviour may vary. 

The approach has some limitations. Primarily, the optimisation of multiple hyperparameters in a single run. We have focused our study on the impact of the discount factor as key element of the Bellman Equation however, there are other hyperparameters that may affect the final system performance. Further work will involve the gradual lifting of these restrictions by allowing the tuning of multiple hyperparameters using different threads or timelines. Another limitation of the approach is the definition of stability on the system which is strongly related to the threshold of stability and time-window length. This could be problematic in situations when the $\mathcal{R}$ is noisy which could produce that the system never gets into a stable condition. This could be tackled by analysing different criteria for stability such as Z-score~\cite{brase2013understanding} and the absolute deviation around the median~\cite{leys2013detecting}. Another approach could be introducing a patience time, e.g. if the system has not entered on a stable condition for X episodes, force it to explore another $\lambda$.

\section{Related Work}
\label{relatedWork}

HPO in the RL traditionally used a Delta-Bar-Delta method as incremental algorithms to tune parameters~\cite{mahmood2012tuning}. However, this method and its variations were limited to linear supervised learning. The recent movement is to combine incremental Delta-Bar-Delta method and Temporal-difference learning ~\cite{young2018metatrace}. Those methods can not make tuning hyperparameters online and allow the algorithm to more robustly adjust to non-stationarity in a problem at the same time. 
A variety of techniques exist to combat this recently— most notably use of a large experience replay buffers or the use of multiple parallel actors. These techniques come at the cost of moving away from the online RL problem as it is traditionally formulated. More sophisticated approaches include Self-Tuning Actor Critic (STAC)~\cite{zahavy2020self} and Sequential Model-based Bayesian optimisation (SMBO)~\cite{feurer2019hyperparameter}. However both methods ignored a crucial issue for RL: the possible non-stationarity of the RL problem induces a possible non-stationarity of the hyperparameters. Thereby, at various stages of the learning process, various hyperparameter settings might be required to behave optimally \cite{zhang2021importance}. Furthermore, these approaches bases their functionality on multiple trials, thus multiple agent’s lifetimes different from the present work that focuses on HPO in a single lifetime.


Moreover, CEP can bring some advantages to ML approaches. They have been used together in fields such as the financial sector ~\cite{luong_open_2020}, cybersecurity~\cite{roldan_integrating_2020} and Internet of Things~\cite{ortiz_real-time_2019}. More particularly, CEP has been used to  preprocess the stream of data that will be provided to the ML classifiers for training and predictive calculations~\cite{luong_open_2020}. More evolved architectures include the use of ML to find and set event patterns for the detection of complex events, thus automatising the setup stage of a CEP system~\cite{mehdiyev_determination_2015}. Even more, some architectures have been developed to automatically update their event patterns using ML~\cite{sun_automatic_2020}. ML and CEP are also combined to provide dynamic fault-tolerance support~\cite{power_providing_2019}.

Recently, CEP has been also integrated with TM to support both service monitoring and explainable reinforcement learning. Specifically, in~\cite{parra-ullauri_event-driven_2021}, an architecture based on CEP and TM is proposed for runtime monitoring of comprehensive data streams. This architecture promptly reacts to events and analyses the historic behaviour of a system. In~\cite{parra-ullauri_event-driven_2021}, a configurable architecture combining CEP and RL allows to keep track of a system's reasoning over time, to extract on-demand history-aware explanations, to automatically detect situations of interest and to real-time filter the relevant points in time to be stored in a TGDB.

\section{Concluding remarks and future work}
\label{conclusion}
Hyperparameter tuning is an omnipresent problem in RL as it is a key element for obtaining the state-of-the-art performance. This paper proposes to tackle this issue by integrating CEP and TMs. We investigated new ways to monitor software agents to explore their environment and pruning algorithms by automatically updating hyperparameters using feedback based on the RL agents historical behaviour. In order to test the feasibility of the approach, we conducted several experiments comparing the performance of a DQN case study using the proposed approach and different traditional HPO techniques. 

The encouraging results show that the proposed framework combining CEP and TMs and implementing the history-aware epsilon-greedy logic significantly improved performance compared to traditional HPO approaches, in terms of reward values and learning speed. Furthermore, the outcomes from the monitoring process produce interpretable results easy for a human to understand and act upon. We have shown how some SE paradigms can be exploited for the benefit of RL and can be further used for creating accountability of RL systems. We believe that the SE and ML communities should work together to solve the critical challenges of assuring the quality of ML/RL and software systems in general.

Future work will include the study of the points mentioned in Section~\ref{discussion} regarding the the limitations of the approach. Further experiments will also be conducted to analyse the performance of the approach with other RL methods. Similarly, we will benchmark the proposed approach with more sophisticated HPO approaches as the ones mentioned in Section~\ref{relatedWork}.

The feedback obtained from the proposed framework can be further exploited for example for safe early stopping~\cite{khamaru2022instance}. Moreover, by choosing a set of high-level operations from hyperparameter tuning to algorithm selection set, to guide an agent to perform various tasks, like remembering history, comparing and contrasting current and past inputs, and using learning methods to change its own learning methods, the proposed approach can be considered a first step towards Meta-Learning~\cite{vanschoren2019meta}. Finally, the obtained knowledge can be useful to convey this information for different stakeholders apart from proving feedback which can be another research direction.

\bibliographystyle{ACM-Reference-Format}
\bibliography{references}


\begin{thebibliography}{31}


\ifx \showCODEN    \undefined \def \showCODEN     #1{\unskip}     \fi
\ifx \showDOI      \undefined \def \showDOI       #1{#1}\fi
\ifx \showISBNx    \undefined \def \showISBNx     #1{\unskip}     \fi
\ifx \showISBNxiii \undefined \def \showISBNxiii  #1{\unskip}     \fi
\ifx \showISSN     \undefined \def \showISSN      #1{\unskip}     \fi
\ifx \showLCCN     \undefined \def \showLCCN      #1{\unskip}     \fi
\ifx \shownote     \undefined \def \shownote      #1{#1}          \fi
\ifx \showarticletitle \undefined \def \showarticletitle #1{#1}   \fi
\ifx \showURL      \undefined \def \showURL       {\relax}        \fi
\providecommand\bibfield[2]{#2}
\providecommand\bibinfo[2]{#2}
\providecommand\natexlab[1]{#1}
\providecommand\showeprint[2][]{arXiv:#2}

\bibitem[Akiba et~al\mbox{.}(2019)]%
        {akiba2019optuna}
\bibfield{author}{\bibinfo{person}{Takuya Akiba}, \bibinfo{person}{Shotaro
  Sano}, \bibinfo{person}{Toshihiko Yanase}, \bibinfo{person}{Takeru Ohta},
  {and} \bibinfo{person}{Masanori Koyama}.} \bibinfo{year}{2019}\natexlab{}.
\newblock \showarticletitle{Optuna: A next-generation hyperparameter
  optimization framework}. In \bibinfo{booktitle}{\emph{Proceedings of the 25th
  ACM SIGKDD international conference on knowledge discovery \& data mining}}.
  \bibinfo{pages}{2623--2631}.
\newblock


\bibitem[Brase and Brase(2013)]%
        {brase2013understanding}
\bibfield{author}{\bibinfo{person}{Charles~Henry Brase} {and}
  \bibinfo{person}{Corrinne~Pellillo Brase}.} \bibinfo{year}{2013}\natexlab{}.
\newblock \bibinfo{booktitle}{\emph{Understanding basic statistics}}.
\newblock \bibinfo{publisher}{Brooks/Cole Cengage Learning}.
\newblock


\bibitem[Esling and Agon(2012)]%
        {esling_time-series_2012}
\bibfield{author}{\bibinfo{person}{Philippe Esling} {and}
  \bibinfo{person}{Carlos Agon}.} \bibinfo{year}{2012}\natexlab{}.
\newblock \showarticletitle{Time-series data mining}.
\newblock \bibinfo{journal}{\emph{ACM CSUR}} \bibinfo{volume}{45},
  \bibinfo{number}{1} (\bibinfo{year}{2012}).
\newblock


\bibitem[Fernandez and Caarls(2018)]%
        {fernandez2018parameters}
\bibfield{author}{\bibinfo{person}{Franklin~Carde{\~n}oso Fernandez} {and}
  \bibinfo{person}{Wouter Caarls}.} \bibinfo{year}{2018}\natexlab{}.
\newblock \showarticletitle{Parameters tuning and optimization for
  reinforcement learning algorithms using evolutionary computing}. In
  \bibinfo{booktitle}{\emph{2018 International Conference on Information
  Systems and Computer Science}}.
\newblock


\bibitem[Feurer and Hutter(2019)]%
        {feurer2019hyperparameter}
\bibfield{author}{\bibinfo{person}{Matthias Feurer} {and}
  \bibinfo{person}{Frank Hutter}.} \bibinfo{year}{2019}\natexlab{}.
\newblock \showarticletitle{Hyperparameter optimization}.
\newblock In \bibinfo{booktitle}{\emph{Automated machine learning}}.
\newblock


\bibitem[G{\'o}mez et~al\mbox{.}(2018)]%
        {gomez2018temporalemf}
\bibfield{author}{\bibinfo{person}{Abel G{\'o}mez}, \bibinfo{person}{Jordi
  Cabot}, {and} \bibinfo{person}{Manuel Wimmer}.}
  \bibinfo{year}{2018}\natexlab{}.
\newblock \showarticletitle{{TemporalEMF}: A temporal metamodeling framework}.
  In \bibinfo{booktitle}{\emph{International Conference on Conceptual
  Modeling}}.
\newblock


\bibitem[Hartmann et~al\mbox{.}(2017)]%
        {greycat}
\bibfield{author}{\bibinfo{person}{Thomas Hartmann}, \bibinfo{person}{Francois
  Fouquet}, {et~al\mbox{.}}} \bibinfo{year}{2017}\natexlab{}.
\newblock \showarticletitle{Analyzing {Complex} {Data} in {Motion} at {Scale}
  with {Temporal} {Graphs}}. In \bibinfo{booktitle}{\emph{Proceedings of
  SEKE'17}}.
\newblock


\bibitem[Jomaa et~al\mbox{.}(2019)]%
        {jomaa2019hyp}
\bibfield{author}{\bibinfo{person}{Hadi~S Jomaa}, \bibinfo{person}{Josif
  Grabocka}, {and} \bibinfo{person}{Lars Schmidt-Thieme}.}
  \bibinfo{year}{2019}\natexlab{}.
\newblock \showarticletitle{Hyp-rl: Hyperparameter optimization by
  reinforcement learning}.
\newblock \bibinfo{journal}{\emph{arXiv preprint arXiv:1906.11527}}
  (\bibinfo{year}{2019}).
\newblock


\bibitem[Khamaru et~al\mbox{.}(2022)]%
        {khamaru2022instance}
\bibfield{author}{\bibinfo{person}{Koulik Khamaru}, \bibinfo{person}{Eric Xia},
  \bibinfo{person}{Martin~J Wainwright}, {and} \bibinfo{person}{Michael~I
  Jordan}.} \bibinfo{year}{2022}\natexlab{}.
\newblock \showarticletitle{Instance-Dependent Confidence and Early Stopping
  for Reinforcement Learning}.
\newblock \bibinfo{journal}{\emph{arXiv preprint arXiv:2201.08536}}
  (\bibinfo{year}{2022}).
\newblock


\bibitem[Klar et~al\mbox{.}(1992)]%
        {klar1992tools}
\bibfield{author}{\bibinfo{person}{Rainer Klar}, \bibinfo{person}{Andreas
  Quick}, {and} \bibinfo{person}{Franz Soetz}.}
  \bibinfo{year}{1992}\natexlab{}.
\newblock \showarticletitle{Tools for a Model-driven Instrumentation for
  Monitoring}. In \bibinfo{booktitle}{\emph{Proceedings of the 5th
  International Conference on Modelling Techniques and Tools for Computer
  Performance Evaluation}}.
\newblock


\bibitem[Leys et~al\mbox{.}(2013)]%
        {leys2013detecting}
\bibfield{author}{\bibinfo{person}{Christophe Leys},
  \bibinfo{person}{Christophe Ley}, \bibinfo{person}{Olivier Klein},
  \bibinfo{person}{Philippe Bernard}, {and} \bibinfo{person}{Laurent Licata}.}
  \bibinfo{year}{2013}\natexlab{}.
\newblock \showarticletitle{Detecting outliers: Do not use standard deviation
  around the mean, use absolute deviation around the median}.
\newblock \bibinfo{journal}{\emph{Journal of experimental social psychology}}
  (\bibinfo{year}{2013}).
\newblock


\bibitem[Luckham and Frasca(1998)]%
        {luckham1998complex}
\bibfield{author}{\bibinfo{person}{David~C Luckham} {and}
  \bibinfo{person}{Brian Frasca}.} \bibinfo{year}{1998}\natexlab{}.
\newblock \showarticletitle{Complex event processing in distributed systems}.
\newblock \bibinfo{journal}{\emph{Computer Systems Laboratory Technical Report
  CSL-TR-98-754. Stanford University, Stanford}} (\bibinfo{year}{1998}).
\newblock


\bibitem[Luong et~al\mbox{.}(2020)]%
        {luong_open_2020}
\bibfield{author}{\bibinfo{person}{Nhan Nathan~Tri Luong},
  \bibinfo{person}{Zoran Milosevic}, \bibinfo{person}{Andrew Berry}, {and}
  \bibinfo{person}{Fethi Rabhi}.} \bibinfo{year}{2020}\natexlab{}.
\newblock \showarticletitle{An open architecture for complex event processing
  with machine learning}. In \bibinfo{booktitle}{\emph{2020 {IEEE} 24th
  {International} {Enterprise} {Distributed} {Object} {Computing}
  {Conference}}}.
\newblock


\bibitem[Mahmood et~al\mbox{.}(2012)]%
        {mahmood2012tuning}
\bibfield{author}{\bibinfo{person}{Ashique~Rupam Mahmood},
  \bibinfo{person}{Richard~S Sutton}, \bibinfo{person}{Thomas Degris}, {and}
  \bibinfo{person}{Patrick~M Pilarski}.} \bibinfo{year}{2012}\natexlab{}.
\newblock \showarticletitle{Tuning-free step-size adaptation}. In
  \bibinfo{booktitle}{\emph{2012 IEEE International Conference on Acoustics,
  Speech and Signal Processing}}.
\newblock


\bibitem[Mazak et~al\mbox{.}(2020)]%
        {mazak2020temporal}
\bibfield{author}{\bibinfo{person}{Alexandra Mazak}, \bibinfo{person}{Sabine
  Wolny}, \bibinfo{person}{Abel G{\'o}mez}, \bibinfo{person}{Jordi Cabot},
  \bibinfo{person}{Manuel Wimmer}, {and} \bibinfo{person}{Gerti Kappel}.}
  \bibinfo{year}{2020}\natexlab{}.
\newblock \showarticletitle{Temporal models on time series databases}.
\newblock \bibinfo{journal}{\emph{J. Object Technol}} (\bibinfo{year}{2020}).
\newblock


\bibitem[Mehdiyev et~al\mbox{.}(2015)]%
        {mehdiyev_determination_2015}
\bibfield{author}{\bibinfo{person}{Nijat Mehdiyev}, \bibinfo{person}{Julian
  Krumeich}, \bibinfo{person}{David Enke}, \bibinfo{person}{Dirk Werth}, {and}
  \bibinfo{person}{Peter Loos}.} \bibinfo{year}{2015}\natexlab{}.
\newblock \showarticletitle{Determination of {Rule} {Patterns} in {Complex}
  {Event} {Processing} {Using} {Machine} {Learning} {Techniques}}.
\newblock \bibinfo{journal}{\emph{Procedia Computer Science}}
  \bibinfo{volume}{61} (\bibinfo{year}{2015}).
\newblock


\bibitem[Mnih et~al\mbox{.}(2015)]%
        {mnih2015human}
\bibfield{author}{\bibinfo{person}{Volodymyr Mnih}, \bibinfo{person}{Koray
  Kavukcuoglu}, \bibinfo{person}{David Silver}, \bibinfo{person}{Andrei~A
  Rusu}, \bibinfo{person}{Joel Veness}, \bibinfo{person}{Marc~G Bellemare},
  \bibinfo{person}{Alex Graves}, \bibinfo{person}{Martin Riedmiller},
  \bibinfo{person}{Andreas~K Fidjeland}, \bibinfo{person}{Georg Ostrovski},
  {et~al\mbox{.}}} \bibinfo{year}{2015}\natexlab{}.
\newblock \showarticletitle{Human-level control through deep reinforcement
  learning}.
\newblock \bibinfo{journal}{\emph{nature}} \bibinfo{number}{7540}
  (\bibinfo{year}{2015}).
\newblock


\bibitem[Moser et~al\mbox{.}(2010)]%
        {moser2010event}
\bibfield{author}{\bibinfo{person}{Oliver Moser}, \bibinfo{person}{Florian
  Rosenberg}, {and} \bibinfo{person}{Schahram Dustdar}.}
  \bibinfo{year}{2010}\natexlab{}.
\newblock \showarticletitle{Event driven monitoring for service composition
  infrastructures}. In \bibinfo{booktitle}{\emph{International Conference on
  Web Information Systems Engineering}}.
\newblock


\bibitem[Ortiz et~al\mbox{.}(2019)]%
        {ortiz_real-time_2019}
\bibfield{author}{\bibinfo{person}{Guadalupe Ortiz},
  \bibinfo{person}{Jose~Antonio Caravaca}, \bibinfo{person}{Alfonso Garcia-de
  Prado}, \bibinfo{person}{Francisco Chavez de~la O}, {and}
  \bibinfo{person}{Juan Boubeta-Puig}.} \bibinfo{year}{2019}\natexlab{}.
\newblock \showarticletitle{Real-{Time} {Context}-{Aware} {Microservice}
  {Architecture} for {Predictive} {Analytics} and {Smart} {Decision}-{Making}}.
\newblock \bibinfo{journal}{\emph{IEEE Access}} (\bibinfo{year}{2019}).
\newblock


\bibitem[Parra-Ullauri et~al\mbox{.}(2021a)]%
        {parra2021towards}
\bibfield{author}{\bibinfo{person}{Juan~Marcelo Parra-Ullauri},
  \bibinfo{person}{Antonio Garc{\'\i}a-Dom{\'\i}nguez}, \bibinfo{person}{Juan
  Boubeta-Puig}, \bibinfo{person}{Nelly Bencomo}, {and}
  \bibinfo{person}{Guadalupe Ortiz}.} \bibinfo{year}{2021}\natexlab{a}.
\newblock \showarticletitle{Towards an architecture integrating complex event
  processing and temporal graphs for service monitoring}. In
  \bibinfo{booktitle}{\emph{Proceedings of the 36th Annual ACM Symposium on
  Applied Computing}}. \bibinfo{pages}{427--435}.
\newblock


\bibitem[Parra-Ullauri et~al\mbox{.}(2021b)]%
        {parra-ullauri_event-driven_2021}
\bibfield{author}{\bibinfo{person}{Juan~Marcelo Parra-Ullauri},
  \bibinfo{person}{Antonio García-Domínguez}, \bibinfo{person}{Nelly
  Bencomo}, \bibinfo{person}{Changgang Zheng}, \bibinfo{person}{Chen Zhen},
  \bibinfo{person}{Juan Boubeta-Puig}, \bibinfo{person}{Guadalupe Ortiz}, {and}
  \bibinfo{person}{Shufan Yang}.} \bibinfo{year}{2021}\natexlab{b}.
\newblock \showarticletitle{Event-driven temporal models for explanations -
  {ETeMoX}: explaining reinforcement learning}.
\newblock \bibinfo{journal}{\emph{Software and Systems Modeling}}
  (\bibinfo{year}{2021}).
\newblock


\bibitem[Power and Kotonya(2019)]%
        {power_providing_2019}
\bibfield{author}{\bibinfo{person}{Alexander Power} {and}
  \bibinfo{person}{Gerald Kotonya}.} \bibinfo{year}{2019}\natexlab{}.
\newblock \showarticletitle{Providing {Fault} {Tolerance} via {Complex} {Event}
  {Processing} and {Machine} {Learning} for {IoT} {Systems}}. In
  \bibinfo{booktitle}{\emph{Proceedings of the 9th {International} {Conference}
  on the {Internet} of {Things}}} \emph{(\bibinfo{series}{{IoT} 2019})}.
  \bibinfo{publisher}{Association for Computing Machinery},
  \bibinfo{address}{New York, NY, USA}.
\newblock


\bibitem[Rabiser et~al\mbox{.}(2017)]%
        {rabiser2017comparison}
\bibfield{author}{\bibinfo{person}{Rick Rabiser}, \bibinfo{person}{Sam Guinea},
  \bibinfo{person}{Michael Vierhauser}, \bibinfo{person}{Luciano Baresi}, {and}
  \bibinfo{person}{Paul Gr{\"u}nbacher}.} \bibinfo{year}{2017}\natexlab{}.
\newblock \showarticletitle{A comparison framework for runtime monitoring
  approaches}.
\newblock \bibinfo{journal}{\emph{Journal of Systems and Software}}
  (\bibinfo{year}{2017}).
\newblock


\bibitem[Roldán et~al\mbox{.}(2020)]%
        {roldan_integrating_2020}
\bibfield{author}{\bibinfo{person}{José Roldán}, \bibinfo{person}{Juan
  Boubeta-Puig}, \bibinfo{person}{José~Luis Martínez}, {and}
  \bibinfo{person}{Guadalupe Ortiz}.} \bibinfo{year}{2020}\natexlab{}.
\newblock \showarticletitle{Integrating {Complex} {Event} {Processing} and
  {Machine} {Learning}: an {Intelligent} {Architecture} for {Detecting} {IoT}
  {Security} {Attacks}}.
\newblock \bibinfo{journal}{\emph{Expert Systems with Applications}}
  \bibinfo{volume}{149} (\bibinfo{year}{2020}).
\newblock


\bibitem[Sun et~al\mbox{.}(2020)]%
        {sun_automatic_2020}
\bibfield{author}{\bibinfo{person}{Yunhao Sun}, \bibinfo{person}{Guanyu Li},
  {and} \bibinfo{person}{Bo Ning}.} \bibinfo{year}{2020}\natexlab{}.
\newblock \showarticletitle{Automatic {Rule} {Updating} based on {Machine}
  {Learning} in {Complex} {Event} {Processing}}. In
  \bibinfo{booktitle}{\emph{2020 {IEEE} 40th {International} {Conference} on
  {Distributed} {Computing} {Systems}}}.
\newblock


\bibitem[Sutton and Barto(2018)]%
        {0c}
\bibfield{author}{\bibinfo{person}{Richard~S Sutton} {and}
  \bibinfo{person}{Andrew~G Barto}.} \bibinfo{year}{2018}\natexlab{}.
\newblock \bibinfo{booktitle}{\emph{Reinforcement learning: An introduction}}.
\newblock \bibinfo{publisher}{MIT press}.
\newblock


\bibitem[Vanschoren(2019)]%
        {vanschoren2019meta}
\bibfield{author}{\bibinfo{person}{Joaquin Vanschoren}.}
  \bibinfo{year}{2019}\natexlab{}.
\newblock \showarticletitle{Meta-learning}.
\newblock In \bibinfo{booktitle}{\emph{Automated Machine Learning}}.
\newblock


\bibitem[Young et~al\mbox{.}(2018)]%
        {young2018metatrace}
\bibfield{author}{\bibinfo{person}{Kenny Young}, \bibinfo{person}{Baoxiang
  Wang}, {and} \bibinfo{person}{Matthew~E Taylor}.}
  \bibinfo{year}{2018}\natexlab{}.
\newblock \showarticletitle{Metatrace: Online step-size tuning by meta-gradient
  descent for reinforcement learning control}.
\newblock \bibinfo{journal}{\emph{arXiv preprint arXiv:1805.04514}}
  (\bibinfo{year}{2018}).
\newblock


\bibitem[Zahavy et~al\mbox{.}(2020)]%
        {zahavy2020self}
\bibfield{author}{\bibinfo{person}{Tom Zahavy}, \bibinfo{person}{Zhongwen Xu},
  \bibinfo{person}{Vivek Veeriah}, \bibinfo{person}{Matteo Hessel},
  \bibinfo{person}{Junhyuk Oh}, \bibinfo{person}{Hado van Hasselt},
  \bibinfo{person}{David Silver}, {and} \bibinfo{person}{Satinder Singh}.}
  \bibinfo{year}{2020}\natexlab{}.
\newblock \showarticletitle{Self-tuning deep reinforcement learning}.
\newblock \bibinfo{journal}{\emph{arXiv preprint arXiv:2002.12928}}
  (\bibinfo{year}{2020}).
\newblock


\bibitem[Zhang et~al\mbox{.}(2021)]%
        {zhang2021importance}
\bibfield{author}{\bibinfo{person}{Baohe Zhang}, \bibinfo{person}{Raghu Rajan},
  \bibinfo{person}{Luis Pineda}, \bibinfo{person}{Nathan Lambert},
  \bibinfo{person}{Andr{\'e} Biedenkapp}, \bibinfo{person}{Kurtland Chua},
  \bibinfo{person}{Frank Hutter}, {and} \bibinfo{person}{Roberto Calandra}.}
  \bibinfo{year}{2021}\natexlab{}.
\newblock \showarticletitle{On the importance of hyperparameter optimization
  for model-based reinforcement learning}. In
  \bibinfo{booktitle}{\emph{International Conference on Artificial Intelligence
  and Statistics}}.
\newblock


\bibitem[Zheng et~al\mbox{.}(2021)]%
        {zheng2021reward}
\bibfield{author}{\bibinfo{person}{Changgang Zheng}, \bibinfo{person}{Shufan
  Yang}, \bibinfo{person}{Juan~Marcelo Parra-Ullauri}, \bibinfo{person}{Antonio
  Garcia-Dominguez}, {and} \bibinfo{person}{Nelly Bencomo}.}
  \bibinfo{year}{2021}\natexlab{}.
\newblock \showarticletitle{Reward-reinforced generative adversarial networks
  for multi-agent systems}.
\newblock \bibinfo{journal}{\emph{IEEE Transactions on Emerging Topics in
  Computational Intelligence}} (\bibinfo{year}{2021}).
\newblock


\end{thebibliography}

\end{document}